\documentclass[conference]{IEEEtran}
\usepackage{cite}
\usepackage{amsmath,amssymb,amsfonts, autobreak, bm}
\usepackage{algorithmic}
\usepackage{graphicx}
\usepackage{textcomp}
\usepackage{subcaption}
\usepackage{listings}
\usepackage{mathtools}
\usepackage{xcolor}
\usepackage{relsize}
\usepackage{dblfloatfix}
\usepackage{color, colortbl}
\usepackage{url}

\usepackage{breakurl}
\definecolor{Gray}{gray}{0.9}
\def\BibTeX{{\rm B\kern-.05em{\sc i\kern-.025em b}\kern-.08em
		T\kern-.1667em\lower.7ex\hbox{E}\kern-.125emX}}

\begin{document}

\title{Adversarial Networks and Machine Learning for File Classification}
\author{\IEEEauthorblockN{Ken St. Germain\textsuperscript{1}, Josh Angichiodo}
	\IEEEauthorblockA{\textit{Department of Cyber Science} \\
		\textit{United States Naval Academy}\\
		Annapolis, MD \\
		\textsuperscript{1}stgermai@usna.edu }
	}



%

\maketitle

\begin{abstract}
Correctly identifying the type of file under examination is a critical part of a forensic investigation.  The file type alone suggests the embedded content, such as a picture, video, manuscript, spreadsheet, etc.  In cases where a system owner might desire to keep their files inaccessible or file type concealed, we propose using an adversarially-trained machine learning neural network to determine a file's true type even if the extension or file header is obfuscated to complicate its discovery.  Our semi-supervised generative adversarial network (SGAN) achieved 97.6\% accuracy in classifying files across 11 different types.  We also compared our network against a traditional standalone neural network and three other machine learning algorithms.  The adversarially-trained network proved to be the most precise file classifier especially in scenarios with few supervised samples available. Our implementation of a file classifier using an SGAN is implemented on GitHub (\url{https://ksaintg.github.io/SGAN-File-Classier/}).
\end{abstract}

\section{Introduction}

Machine learning can be used to determine file types based on a file's byte value distribution.  In this work, we introduce an adversarial learning approach to accurately identify file types regardless of file extension, headers, or footers.  By inspecting the histogram-based distribution of byte values in a file, we can greatly reduce the time and effort expended by subject matter experts during the course of a forensic investigation.

Machine learning algorithms are designed to extract relevant information from data \cite{deisenroth_mathematics_2020}, and the field of deep learning has been shown effective in solving classification problems \cite{goodfellow_deep_2016}.  In this paper we use a generative adversarial network (GAN) to determine the type of file under investigation.  Specifically, we employ a GAN model with semi-supervised learning known as a semi-supervised GAN (SGAN) \cite{odena_semi-supervised_2016} where only a small portion of the training dataset is labeled.

\subsection{Hiding files}

Privacy advocates \cite{fakhoury_know_2014} urge users to protect their private information from criminal interception or unlawful government overreach, and protecting the digital data stored on users' computers, phones, and other devices can include denying physical access or employing encryption.  While encryption has become more commonplace and accessible \cite{peterson_everything_2015}, users desiring more security against cryptographic weaknesses \cite{barenghi_fault_2012}, \cite{mukherjee_principles_2014} may apply additional measures to safeguard their information.

By changing file extensions or removing them altogether, a user can obfuscate the true file type.  While this rudimentary technique applied to a small number of files may not be a challenge to computer forensic investigators, it may be more effective if used across a large body of files composed of varying types.

Many operating systems will select (or suggest) an application to open a file based on the file extension  \cite{noauthor_filename_2006}.  For example, Microsoft Windows will use the file extension, such as {\fontfamily{pcr}\selectfont .docx} to determine the application to open the file.  A file named {\fontfamily{pcr}\selectfont cat.docx} suggests that the file is a document that can be opened by Microsoft Word.  However, users can change the names and extensions of the file to any arbitrary string of characters.  A file originally created as a bitmap file named {\fontfamily{pcr}\selectfont cat.bmp} and renamed to {\fontfamily{pcr}\selectfont cat.docx} will not open and render correctly using Microsoft Word.

There are a variety of reasons to keep the nature of a file unknown to all but the user.  By obfuscating file types, malware developers may hope to evade email filters or anti-virus software \cite{microsoft_blocked_nodate} \cite{collins_dont_2021}. A user engaged in illicit activities may desire to hinder law enforcement by complicating evidence discovery \cite{freeh_impact_1997}.  Whatever the user's motivation, without the correct file extension and absent a brute-force approach, an investigator will require a tool to efficiently discover the appropriate program to open the file.

\subsection{Finding files}

Many 
file types can be determined by examining the file header and footer information, also known as a ``magic number".  The file header is the first few bytes in a file and the footer is the last few bytes in a file.  Depending on file type, the file headers and footers will be of various lengths and have different values.  Many file types will have unique headers and footers, yet some file types will share header and footer values, e.g., {\fontfamily{pcr}\selectfont .xls}, {\fontfamily{pcr}\selectfont .doc}, {\fontfamily{pcr}\selectfont .ppt} \cite{rentz_microsoft_2007}.

File headers and footers can be analyzed through command-line tools that perform a binary or hexadecimal dump, or by using binary or hexadecimal readers/editors to provide insight to the file type.  Alternatively, tools like \textit{Scalpel} \cite{richard_iii_scalpel_2005} search a chunk of data that may contain multiple files, and based on user-configured options, will perform file carving that allows the investigator to see the chunk's number and file types within.  \textit{Scalpel's} configurable options use header and footer values as well as common signatures within a file.  For example, although an {\fontfamily{pcr}\selectfont html} file is plaintext and will not have a header, it will likely include the text string {\fontfamily{pcr}\selectfont <html>}.

Regardless of an investigator's methods, specialized knowledge is required to conclude the type of file under examination.  If the hexadecimal string {\fontfamily{pcr}\selectfont D0 CF 11 E0 A1 B1 1A E1} is found in the header, this could be one of five Microsoft Office file types \cite{rentz_microsoft_2007}.  When several thousand or more files require classification, the time demand on the most experienced investigator greatly increases.

\subsection{Contributions}

This work uses machine learning algorithms trained on extracted file features to identify the type of file under investigation.  We created histograms based on the frequency of byte-values (ranging from zero to 255) to train and then test our machine learning algorithms.  Specifically, our contribution provides:
\begin{itemize}
\item A classifier from a semi-supervised generative adversarial network designed to identify file types
\item Comparison of classifier accuracy with the performance of a  traditionally-trained multi-layer perceptron (MLP) network
\item Comparison and analysis of the neural network method compared to the results from non-neural network machine learning algorithms, specifically Decision Tree, extreme gradient boosting (XGBoost), and k-Nearest Neighbor (kNN)
\end{itemize}

To the best of our knowledge, no other work has used a classifier of an adversarially-trained neural network to conduct file type classification.  We show improved accuracy over previously explored methods can be achieved with reduced expert analysis required to create samples for a training dataset.

This paper provides background and discussion of related works in Section~\ref{background}.  We then discuss our dataset and how we derive our samples for machine learning in Section~\ref{dataset}.  We present our SGAN architecture in Section~\ref{sgan_arch} and discuss other machine learning algorithms in Section~\ref{mach_algs}.  The results of our work are summarized in Section~\ref{results} and we provide our conclusions and future work in Section~\ref{conclusion}.

\section{Background}\label{background}
This section examines previous work in file classification and introduces the SGAN.  We summarize the use of byte values within files to determine file types and we discuss the use of machine learning in file classification.  Finally, we discuss the nature of adversarial networks and examine the SGAN model.  

\subsection{Classification using byte values}

As an alternative to header and footer inspection, McDaniel and Heydari used the binary content of files to identify the type in \cite{mcdaniel_content_2003}.  They used several algorithms based on a byte frequency distribution fingerprint to determine a file type, showing that file classification can be accomplished by comparing a candidate file's byte distribution to the distribution of 120 other files of known type.  The accuracy of their proposed algorithms was just under 96\% when they grouped together {\fontfamily{pcr}\selectfont .acd}, {\fontfamily{pcr}\selectfont .doc}, {\fontfamily{pcr}\selectfont .xls}, and {\fontfamily{pcr}\selectfont .ppt} file types into one class.  When these files were separately classified, the accuracy rate dropped to 85\%.  Based on the binary frequency distribution in \cite{mcdaniel_content_2003}, several authors have extended the research on file classification.  

In \cite{li_fileprints_2005}, Li et al. were able to improve on McDaniel and Heydari's accuracy in \cite{mcdaniel_content_2003} using a centroid-based approach and saw improved accuracy when truncating the files.  Li used the Manhattan Distance for each files' byte distributions to compare files and determine the appropriate classification.  Because of file header similarity, Li created centroid models that combined file types similar to McDaniel's approach in \cite{mcdaniel_content_2003}.  Specifically, there was one model that combined {\fontfamily{pcr}\selectfont .exe} and {\fontfamily{pcr}\selectfont .dll} files into one class, and another model that combined {\fontfamily{pcr}\selectfont .doc}, {\fontfamily{pcr}\selectfont .xls}, and {\fontfamily{pcr}\selectfont .ppt} files together in another class.

Moody and Erbacher introduced the Statistical Analysis Data Identification (SADI) algorithm in \cite{moody_sadi_2008}.  After calculating byte values for each file, a range of statistical information was gathered and subsequently used to determine file types.  The accuracy of SADI had varying success with nine different file types, reaching 76\% accuracy of all file types after initial analysis.  A secondary assessment on file types that previously did not reach greater than 92\% accuracy showed improvement when characteristic patterns were considered.

Using fragments of {\fontfamily{pcr}\selectfont .pdf}, {\fontfamily{pcr}\selectfont .rtf}, and {\fontfamily{pcr}\selectfont .doc} files from a publicly-available dataset \cite{garfinkel_bringing_2009}, Rahmet et al. leveraged longest common sub-sequences to identify file fragments in \cite{rahmat_file_2017}.  The authors' algorithm successfully classified these file fragments with 92.91\% overall accuracy.

Our work extends the efforts discussed here, and we also made use of byte values and the frequency in which they arose in a file.  The byte value distribution was provided to machine learning algorithms, and each file type was classified.  While we also used file types that shared the same header strings and files that did not contain headers, we created models that differentiated the files uniquely instead of choosing to group them together.

\subsection{Machine learning for file classification}

In \cite{amirani_new_2008}, Amirani et al. used principle component analysis (PCA) and neural networks to achieve file classification accuracy of 98.33\% against a pool of six different file types.  The authors used two neural networks: a five-layer MLP network that uses PCA features as the input, and a second three-layer MLP network to conduct file classification.  Each of their six file types were equally represented in the dataset, with 120 files of each type.

Konaray et al. conducted several experiments using a variety of machine learning algorithms in \cite{konaray_detecting_2019}.  The dataset used by Konaray were composed of 13 text-based files (e.g., {\fontfamily{pcr}\selectfont .html}, {\fontfamily{pcr}\selectfont .py}, {\fontfamily{pcr}\selectfont .bat}, etc.).  The authors were able to achieve an accuracy of 97.83\% using the XGBoost algorithm \cite{chen_xgboost_2016}.  

Comparing statistical classification algorithms such as support vector machine (SVM) and kNN with commercially available tools, Gopal et al. showed that machine learning algorithms could outperform commercial products in \cite{gopal_statistical_2011}.  The authors collected byte values for their experiments using an n-gram approach.  They showed that kNN with 1-gram byte values and SVM with 2-gram byte values greatly outperformed commercial tools in terms of accuracy.

Inspired by the efforts in machine learning research, we also hope to improve file classification accuracy.  As we will discuss in Section~\ref{dataset}, the dataset we used provided access to more file types and of a wider variety than those mentioned in the works here.  The present work uses 11 types of files, including some that are solely composed of ASCII characters such as {\fontfamily{pcr}\selectfont .txt} and {\fontfamily{pcr}\selectfont .html}.  In order to further research in this domain, we investigated the SGAN-trained classifier.

\subsection{Semi-supervised GAN}\label{sgan}
Semi-supervised learning requires that only a portion of the training data be labeled.  Semi-supervised learning differs from supervised learning where all training data is labeled, and also unsupervised learning, where no labels exist and the networks must find their own way to organize the data.  Semi-supervised learning is valuable for large training data sets when it would be laborious and time-intensive to manually label each file.  

When training an unsupervised GAN, the discriminator, $\mathcal{D}$, is a two-class classifier that receives authentic samples from the training dataset or spoofed samples created by the generator, $\mathcal{G}$.  The generator uses random variable input to create the fake samples and the parameters in $\mathcal{G}$. The discriminator assigns a probability from zero to one based on its assessment that the sample is fake (0.0) or authentic (1.0).  The value function that describes this relationships from the original work by Goodfellow \cite{goodfellow_generative_2014-1} is given by 
\begin{equation}\label{eq:minimax}
	\begin{split}
\min_G \max_D V(D, G) = &
		 ~\mathbb{E}_{x\sim p_{data}(x)}[\log D(x)] \\
	 +~&\mathbb{E}_{z\sim p_{z}(z)}[\log(1 - D(G(z)))]
	\end{split}
\end{equation}
where $\mathit{D(x)}$ is the probability that $\mathit{x}$ came from the data distribution $\mathit{p_{data}(x)}$ containing authentic training samples, and $\mathit{D(G(z))}$ is the estimate of the probability that the discriminator incorrectly identifies the fake instance as authentic. The generator network attempts to maximize $\mathit{D(G(z))}$, while the discriminator network tries to minimize it.  The generator creates samples, $\mathit{G(z)}$, based on the parameter values in $\mathit{G}$ and the random values $z$ provided to the generator consistent with $\mathit{p_{z}(z)}$.

With semi-supervised learning, a small percentage of the training data is labeled and the discriminator becomes a multi-class classifier.  For $N$ classes, the model requires ${N+1}$ outputs to account for all the authentic classes plus one additional class for the fake generated class. This can be implemented in a variety of ways.  Following Salimans et al. \cite{salimans_improved_2016}, we can build an $N$-class classifier network, $\mathcal{C}$, with output logits ${\{l_1, l_2, \dots, l_N\}}$ prior to a \textit{softmax} activation for $\mathcal{C}$.  The logits vector is used as the input to a single perceptron followed by the \textit{sigmoid} activation function for $\mathcal{D}$.  The \textit{sigmoid} function is given as ${\sigma(z) = \frac{1}{1+e^{-z}}}$, where $z$ is the output value of the discriminator output layer perceptron.  Because $\mathcal{D}$ and $\mathcal{C}$ share the same input and hidden layer weights, both networks act as a single network, $\mathcal{D/C}$, that is updated during backpropagation based on their respective loss functions, $J^{(\mathcal{D})}$ and $J^{(\mathcal{C})}$.  The generator loss function is given by $J^{(\mathcal{G})}$.  


\begin{figure}[b]
	\centering     
	\includegraphics[width=0.48\textwidth]{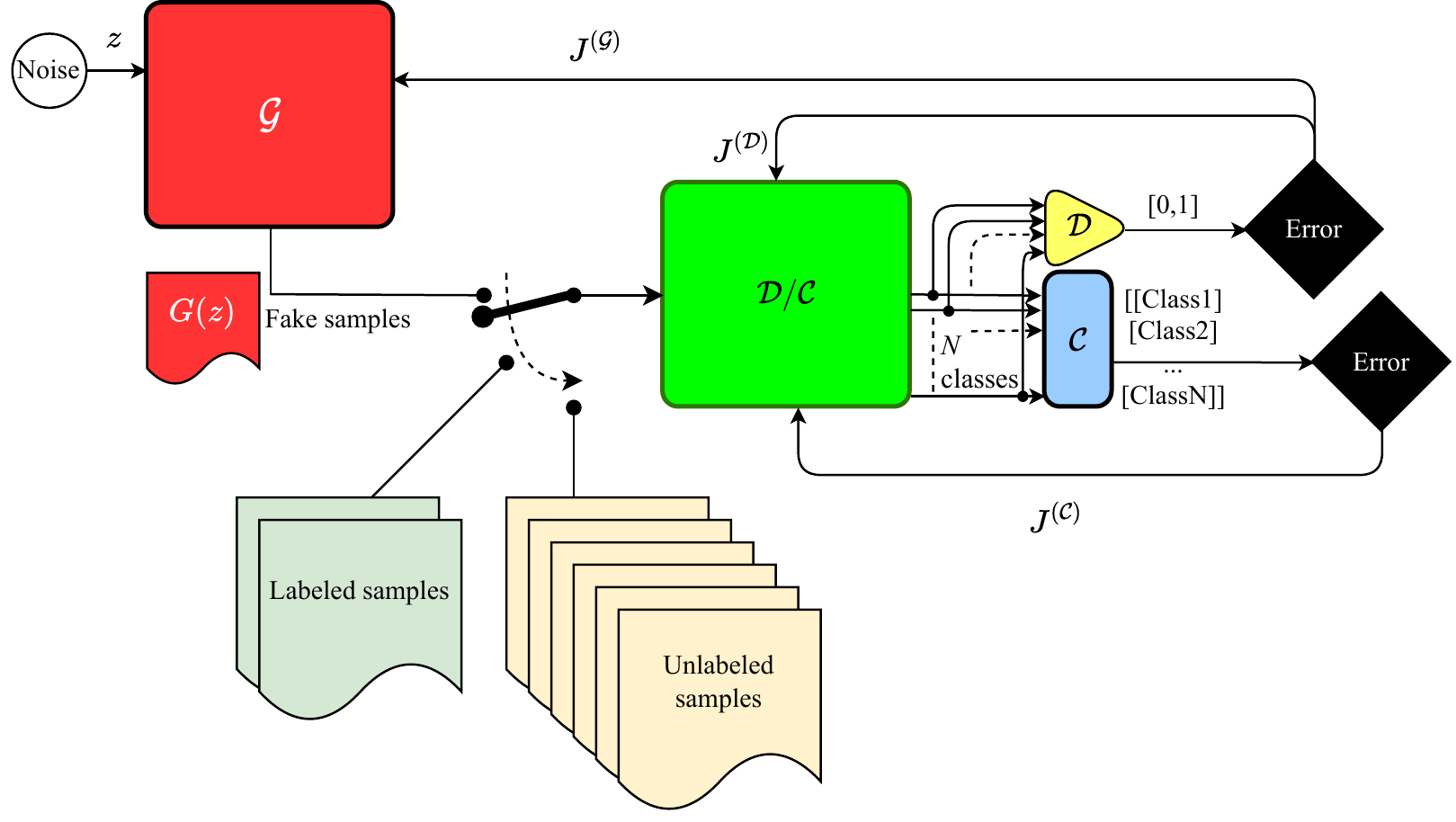}
	\caption{Training a semi-supervised generative adversarial network with N classes.}
	\label{fig:gan}
\end{figure}

Figure~\ref{fig:gan} shows a functional depiction of an SGAN in training.  The training dataset is partially labeled and provided to the $\mathcal{D/C}$ model for classification by $\mathcal{C}$.  The remainder of the training dataset as well as the generated samples from $\mathcal{G}$ are used as input to $\mathcal{D/C}$ for discrimination where $\mathcal{D}$ will predict whether the sample came from the training dataset or if it was created by $\mathcal{G}$.

\section{Dataset}\label{dataset}

\begin{figure*}[]
	\centering     
	\includegraphics[width=0.95\textwidth]{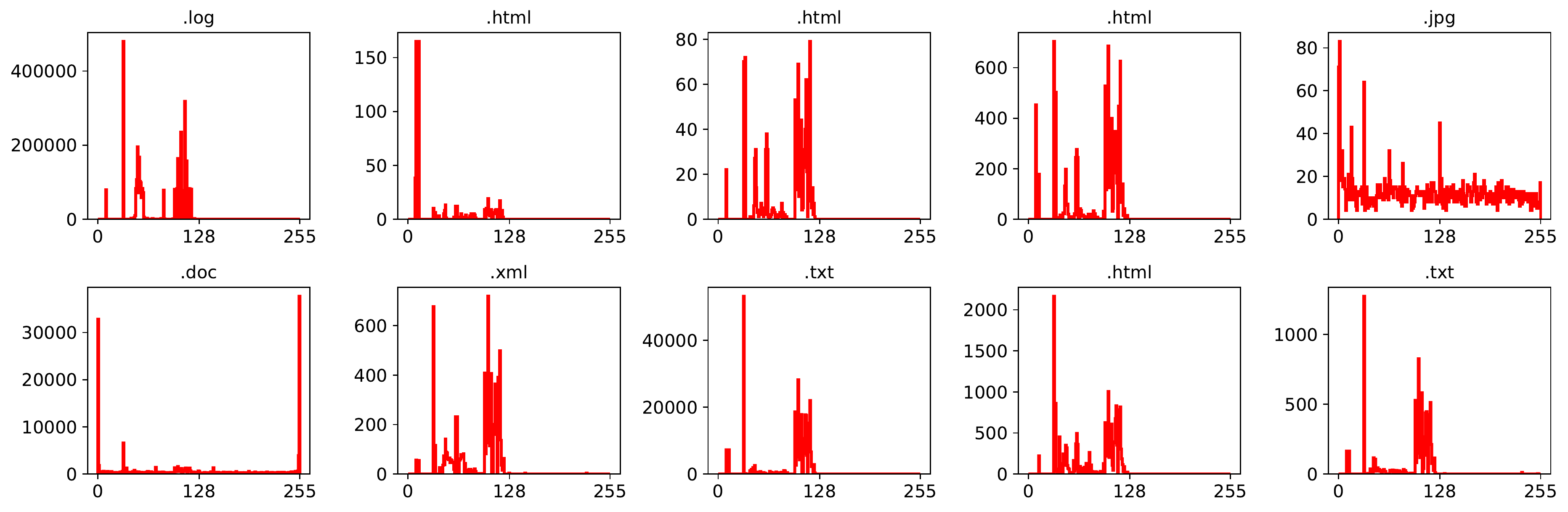}
	\caption{Sample of histograms showing byte value distribution for various files.}
	\label{fig:hist_examples}
\end{figure*}

A dataset containing a variety of different files was desired to ensure we could discern among a range of files.  We used \textit{Govdocs1}, a publicly-available repository of about one million files taken from webservers in the .gov domain \cite{garfinkel_bringing_2009}.  The entire \textit{Govdocs1} corpus consists of 1,000 directories, however we only used the first three folders (000, 001, and 002) creating a total dataset of 2,946 files, totaling 1.56 GB.  We chose to limit the dataset to ensure the processing demands would not require exceptional computational resources.  This work was accomplished using a laptop computer with a 2.60 GHz Intel i7 processor and 32 GB RAM.  Limiting the dataset also allows our work to be easily reproduced.  

The  dataset contained many common file types to include {\fontfamily{pcr}\selectfont .csv}, {\fontfamily{pcr}\selectfont .doc}, {\fontfamily{pcr}\selectfont .gif}, {\fontfamily{pcr}\selectfont .html}, {\fontfamily{pcr}\selectfont .jpg}, {\fontfamily{pcr}\selectfont .pdf}, {\fontfamily{pcr}\selectfont .txt}, {\fontfamily{pcr}\selectfont .xls}, etc.  We noted an unequal distribution of these files such as 28 {\fontfamily{pcr}\selectfont .csv} files, 254 {\fontfamily{pcr}\selectfont .doc} files, and 726 {\fontfamily{pcr}\selectfont .pdf} files.  Unfortunately there were some types that were especially underrepresented, including one {\fontfamily{pcr}\selectfont .gls} file and two {\fontfamily{pcr}\selectfont .java} files.

\subsection{Histograms}

To capture byte value distributions, every file was converted to a histogram.  Each histogram contained 256 bins in the range [0~,~255], representing the decimal value of each byte in the file.  For every bin, the frequency of that decimal value occurring in the file was recorded.  Histogram examples are shown in Figure~\ref{fig:hist_examples}.  In each plot, the bins are shown on the horizontal axis while the frequency value is represented on the vertical axis.

As Figure~\ref{fig:hist_examples} shows, there are differences in the byte distribution between both files of the same type and files of different types, but there are also similarities in different file types such as {\fontfamily{pcr}\selectfont .txt} and {\fontfamily{pcr}\selectfont .html} files.  Machine learning is an appropriate tool to capture the histogram distributions and not only differentiate among the different file types but also group together files of matching type despite varying byte values.

\begin{figure}[b]
	\centering  
	\subfloat[\label{fig:pdf_unscaled}]{\includegraphics[width=0.23\textwidth]{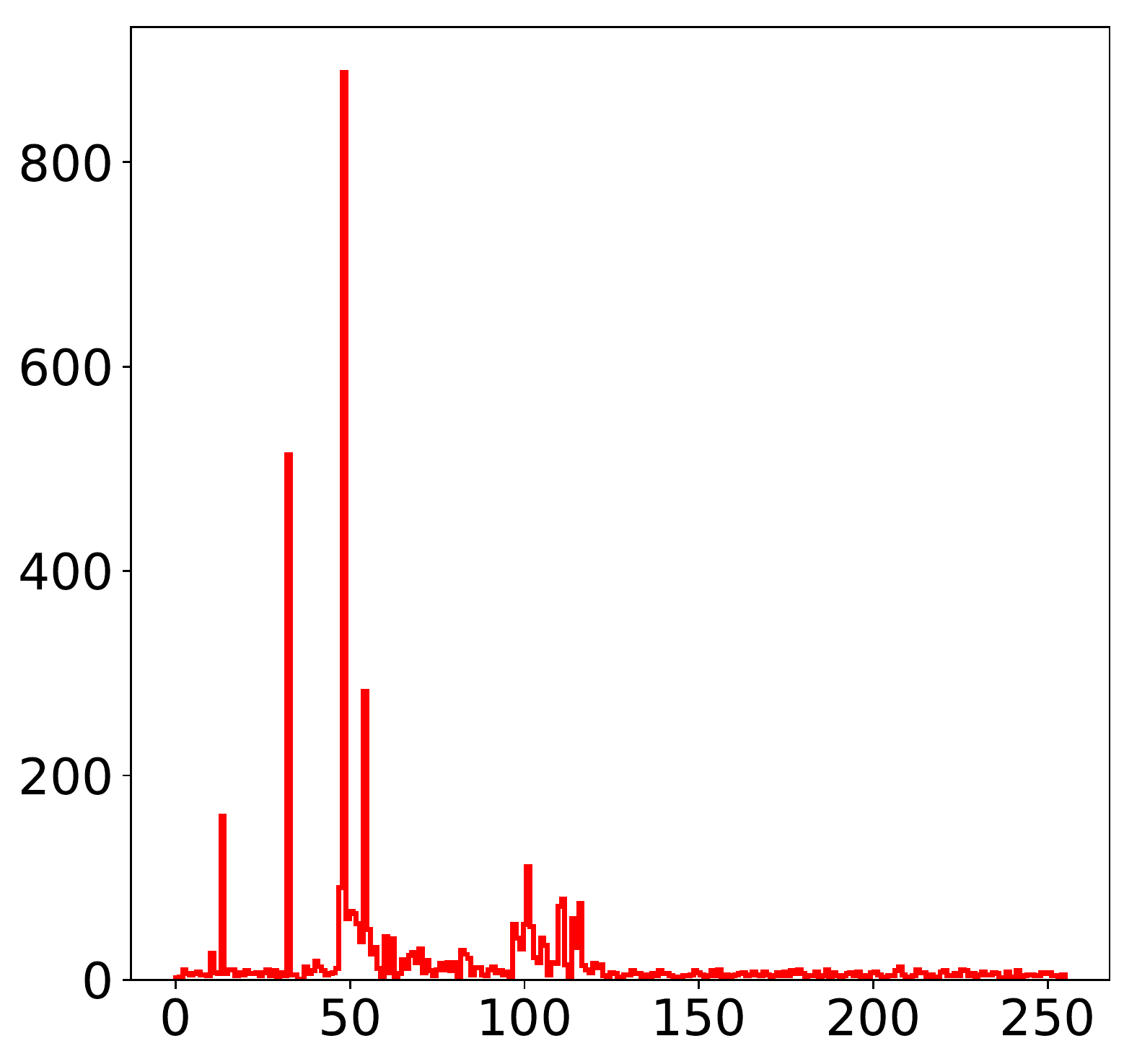}}\hfill
	\subfloat[\label{fig:pdf_scaled}]{\includegraphics[width=0.24\textwidth]{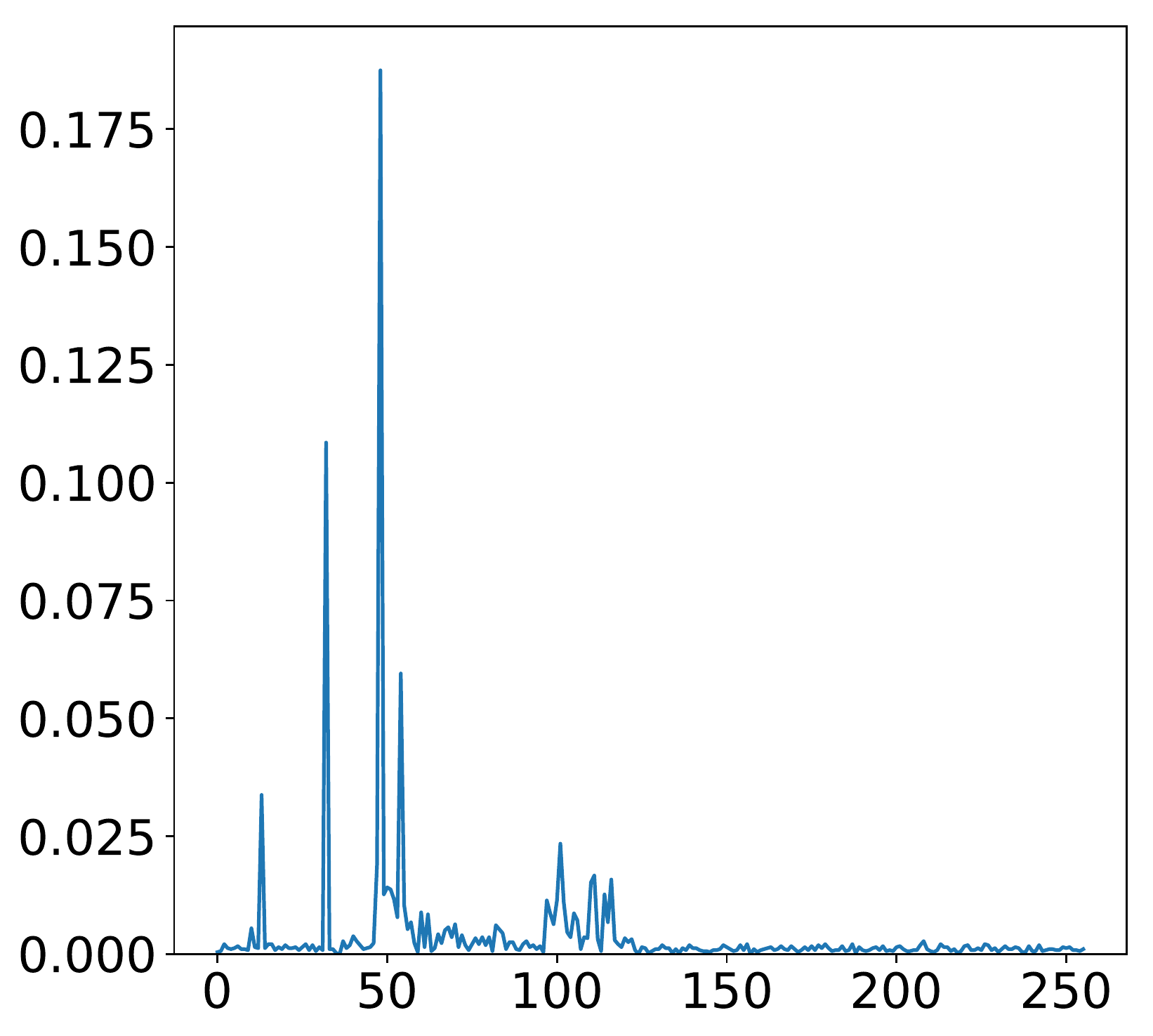}}
	\caption{Example histogram samples showing (a)~an~unscaled {\fontfamily{pcr}\selectfont .pdf} file and (b)~a~normalized {\fontfamily{pcr}\selectfont .pdf} file.}
	\label{fig:pdf}
\end{figure}

\subsection{Samples}

After creating histograms for each file, we then processed the histograms into samples.  To ensure consistency across our samples regardless of file size, we normalized each histogram, scaling each to a cumulative distribution of 1.0.  Figure~\ref{fig:pdf} shows the same {\fontfamily{pcr}\selectfont .pdf} file where Figure~\ref{fig:pdf_unscaled} is the original and Figure~\ref{fig:pdf_scaled} is normalized.

Since insufficient sample sizes for each class can precipitate classification error \cite{raudys_small_1991}, we removed the file types that appeared less than 20 times, representing less than 0.7\% of the total.  There were 14 different file types and 86 total files removed, leaving our dataset with 11 classes and 2860 samples.  Our dataset's composition is shown in Table~\ref{tab:file_types}.  The sample order was then shuffled and finally split into training and testing datasets.  The training dataset used 80\% of the total samples, while the remaining 20\% were reserved for testing.  

\begin{table}[b]
\centering
\caption{Dataset file composition}
\label{tab:file_types}
\resizebox{0.45\textwidth}{!}{%
\begin{tabular}{l|lllllllllll} 
\hline
file type & .csv & .doc & .gif & .html & .jpg & .pdf & .ppt & .ps & .txt & .xls & .xml \\\
samples   & 28   & 254  & 40   & 681   & 229  & 726  & 207  & 40  & 486  & 137  & 32 \\
\hline
\end{tabular}}
\end{table}

\section{SGAN architecture}\label{sgan_arch}

The adversarial competition in the SGAN is a minimax game described by (\ref{eq:minimax}) where the discriminative model attempts to correctly identify authentic training samples from a distribution produced by the scaled histograms representing the dataset files, $p_{data}$, and fake training samples created by the generator.      

While $\mathcal{D}$ and $\mathcal{G}$ adversarially train each other, they learn to improve their individual performances.  Additionally, $\mathcal{C}$ is trained on labeled samples from the training dataset.  Although $\mathcal{C}$ does not directly receive unlabeled authentic or fake samples, the weights of $\mathcal{C}$ are affected by unsupervised training since it shares weights with $\mathcal{D}$ in the $\mathcal{D/C}$ implementation.

The SGAN was implemented using the Python programming language, Keras \cite{chollet_keras_2015} front-end, and Tensorflow \cite{abadi_tensorflow_2015} back-end.  Additionally, Numpy, and Matplotlib Python libraries were used.  The overall SGAN design is summarized in Table~\ref{tab:gan_full}, with a total of 417,271 parameters for the discriminator and the generator, and 304,779 parameters for the classifier.  The file size of the classifier was 3,634 KB.

\begin{table}[]
\centering
\caption{SGAN architecture}
\label{tab:gan_full}
\resizebox{0.45\textwidth}{!}{%
\begin{tabular}{ll|l|l}
\multicolumn{2}{l}{Discriminator/Classifier:}                & \multicolumn{1}{l}{}                      &             \\ 
\cline{2-4}
 & layer                                                     & output size                               & activation  \\ 
\cline{2-4}
 & Input: $x\sim p_{data}(x)$                                & 256                                       &             \\
 & Fully Connected                                           & 512                                       & ReLU        \\
 & Dropout = 0.3                                             &                                           &             \\
 & \begin{tabular}[c]{@{}l@{}}Fully Connected\\\end{tabular} & 256                                       & ReLU        \\
 & Dropout = 0.3                                             &                                           &             \\
 & Fully Connected                                           & 128                                       & ReLU        \\
 & Dropout = 0.3                                             &                                           &             \\
 & Fully Connected                                           & 64                                        & ReLU        \\
 & Dropout = 0.3                                             &                                           &             \\
 & Fully Connected                                           & 11  $l_n = {\{l_1, l_2, \dots, l_{11}\}}$ &             \\
 & Discriminator Output                                      & 1                                         & sigmoid     \\
 & Classifier Output                                         & 11                                        & softmax     \\ 
\cline{2-4}
 & \multicolumn{1}{l}{}                                      & \multicolumn{1}{l}{}                      &             \\
\multicolumn{2}{l}{Generator:}                               & \multicolumn{1}{l}{}                      &             \\ 
\cline{2-4}
 & layer                                                     & output                                    & activation  \\ 
\cline{2-4}
 & Input: $z\sim p_{z}(z)$                                   & 100                                       &             \\
 & Fully Connected                                           & 32                                        & ReLU        \\
 & Dropout = 0.3                                             &                                           &             \\
 & Fully Connected                                           & 64                                        & ReLU        \\
 & Dropout = 0.3                                             &                                           &             \\
 & Fully Connected                                           & 128                                       & ReLU        \\
 & Dropout = 0.3                                             &                                           &             \\
 & Fully Connected                                           & 256                                       & ReLU        \\
 & Output                                                    & 256                                       & sigmoid    
\end{tabular}
}
\end{table}

The discriminator/classifier network, $\mathcal{D/C}$, is a densely or fully connected MLP deep neural network (DNN) with a single input for the file histograms.  Four additional fully connected layers of size 512, 256, 128 and 64 are followed with rectified linear unit (\textit{ReLU}) activation functions.  The \textit{ReLU} function, $g$ is given by ${g(z) = \text{max}(0,z)}$.  The four hidden layers use \textit{Dropout} of 0.3 to prevent overfitting.  Prior to the output layers, a fully connected layer of size~11 is used to capture the number of file types to be classified.  The discriminator output layer of size 1 is fully connected and uses a \textit{sigmoid} activation function to provide values [0.0,~1.0] as discussed in Section~\ref{sgan}.  The classifier output is a \textit{softmax} activation connected to the 11 node layer.  The \textit{softmax} function indicates the most likely class to which the input belongs.  The learning rate for $\mathcal{D/C}$ was 0.0005 using the \textit{Adam} \cite{kingma_adam_2014} optimizer and training was done with batches of 32 samples.


The generator network, $\mathcal{G}$, has a single input with 100~nodes fully connected to the first hidden layer of size~32.  Two additional hidden layers of sizes 64 and 128 are again fully connected using \textit{ReLU} activations.  Finally, a layer of size~256 is connected to the output layer and \textit{sigmoid} activation that ultimately creates the fake histograms samples.  The learning rate for $\mathcal{G}$ was 0.0005 using the \textit{Adam} optimizer.

\section{Machine learning algorithms}\label{mach_algs}

In order to illustrate the SGAN's performance when classifying files, we used additional machine learning algorithms.  We assessed another neural network, the decision trees learning method, the XGBoost algorithm, and the nearest neighbors algorithm.  The same training and testing dataset were used for each machine learning model.  The SGAN was the most complex to train due to using multiple neural networks and no convergence to a global minima.

In terms of structure, the closest model to the SGAN is a supervised learning-based neural network.  We created an MLP network with identical architecture to our SGAN classifier.  The standalone MLP network was trained in a fully supervised manner to accurately select the correct file type based on input.  Both the SGAN and standalone MLP models were trained with a batch size of 32 samples, and training was limited to no more than 300 epochs.  Following training, the best classifiers were selected based on their accuracy against the training dataset.  These classifiers were then evaluated on the test dataset as reported in Section~\ref{results}.

Decision trees are a supervised learning approach that can be used to accomplish multi-class classification \cite{pedregosa_scikit-learn_2011}.  Using the features of the histograms, the decision tree algorithm examines the parametric values in each sample and attempts to accurately classify the file based on a series of decisions based on learned thresholds.

The XGBoost algorithm was implemented as a classifier.  XGBoost is a supervised learning tool that can be used to help us predict the correct file type category.  With multiple classes, the multi-class logistic loss function was used to train the model. 

Finally, the nearest neighbors classification algorithm compares measurements of the input data and training data \cite{pedregosa_scikit-learn_2011}  based on previously stored training information.  The classification result is determined by the number of samples selected, k, with the smallest Euclidean distance among the sample attributes.  We iterated k from one to six to determine the most appropriate number of neighbors to consider when deciding the classification.

\section{Results}\label{results}

\begin{table*}[]
	\centering
	\caption{Classification Accuracy}
	\label{tab:acc_results}
	\resizebox{0.95\textwidth}{!}{%
		\begin{tabular}{|l|l|l|l|l|l|l|l|l|l|l|} 
			\hline
			\begin{tabular}[c]{@{}l@{}}Number of \\supervised samples\end{tabular} & SGAN                                                 & \begin{tabular}[c]{@{}l@{}}Standalone \\ MLP\end{tabular} & Decision Tree                               & XGBoost & kNN, k = 1                                  & kNN, k = 2                                  & kNN, k = 3                                   & kNN, k = 4                                  & kNN, k = 5                                  & kNN, k = 6                                   \\ 
			\hline
			\rowcolor{Gray} 2288                                 & \textbf{0.97552} & 0.96154   & 0.90734   & 0.90384    & 0.88986     & 0.82692     & 0.874126    & 0.83042     & 0.85490      & 0.81293      \\
			1144                                                & \textbf{0.93357} & 0.92132    & 0.86363            & 0.87413        & 0.86713     & 0.79720      & 0.84091     & 0.75350      & 0.81469     & 0.76049      \\
			\rowcolor{Gray} 500                                  & \textbf{0.91783} & 0.9021    & 0.82168            & 0.77972        & 0.84965     & 0.71504     & 0.76573     & 0.62063     & 0.74650      & 0.65734      \\
			100                                                  & \textbf{0.87413} & 0.81469   & 0.48252            & 0.65559        & 0.71504     & 0.44406     & 0.61189     & 0.48427     & 0.52800       & 0.38990       \\
			\rowcolor{Gray}50                                   & \textbf{0.81993} & 0.62062    & 0.26573            & 0.56818        & 0.66084     & 0.38112     & 0.54895     & 0.43007     & 0.30944     & 0.08741      \\
			\hline
	\end{tabular}}
\end{table*}

\begin{figure}[b]
	\centering  
	\subfloat{\includegraphics[width=0.45\textwidth]{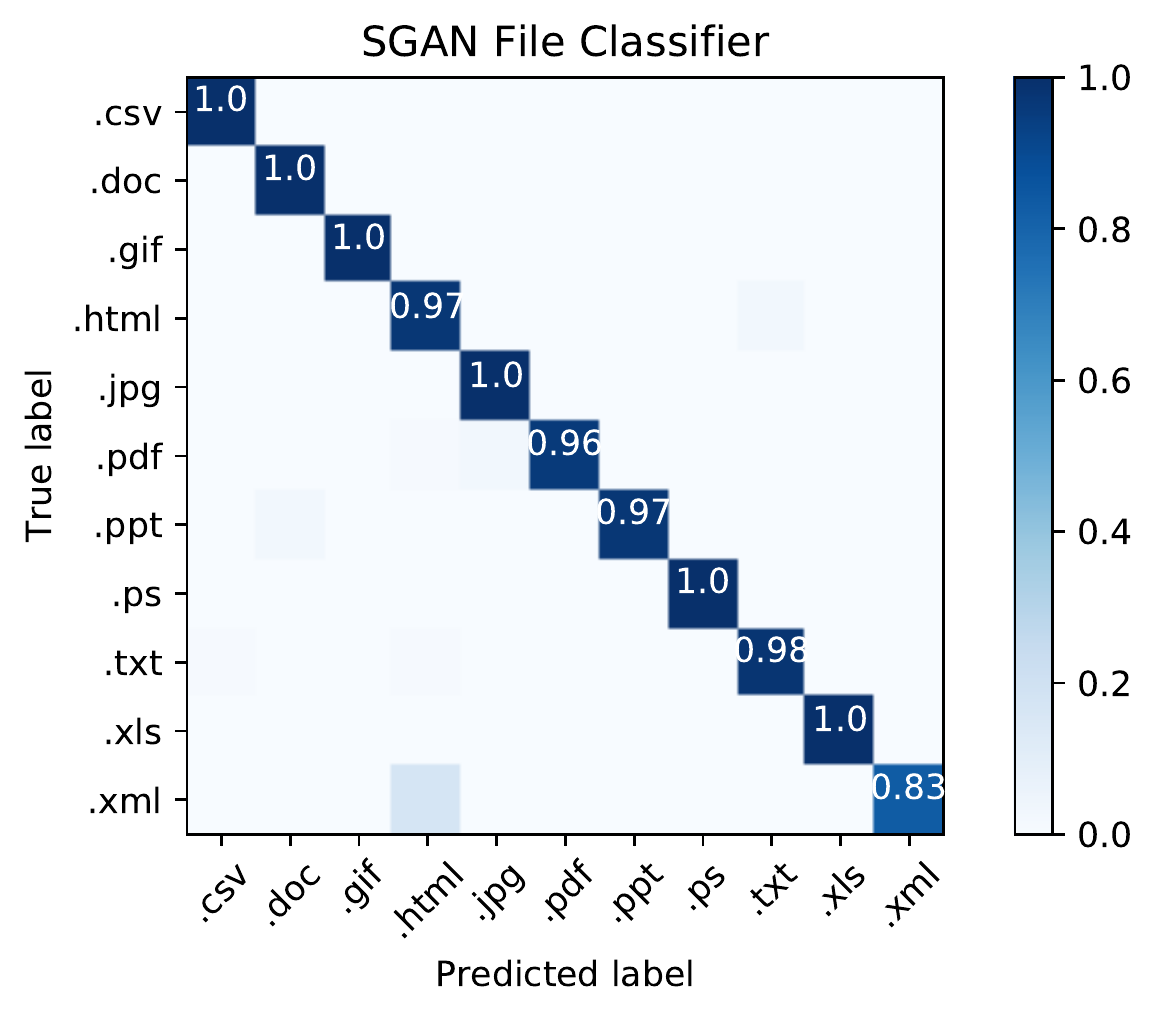}}
	\caption{Confusion matrix for fully-supervised SGAN.}
	\label{fig:full_SGAN}
\end{figure}

Our results are summarized in Table~\ref{tab:acc_results}.  The SGAN was most accurate among all other machine learning algorithms regardless of the number of supervised samples used in training.  When using the entirety of the training data for training the SGAN classifier, we achieved the highest classification performance with the SGAN at 97.552\% accuracy.  Figure~\ref{fig:full_SGAN} shows the confusion matrix of the SGAN when the classifier had access to the entire training data.  We see that the SGAN performed worst at identifying {\fontfamily{pcr}\selectfont .xml} files at 83\% accuracy, confusing them with {\fontfamily{pcr}\selectfont .html} files.  Looking over our dataset, we note that {\fontfamily{pcr}\selectfont .xml} files were among the fewest number of samples available for training.  For some test samples, the SGAN confused {\fontfamily{pcr}\selectfont .ppt} files with {\fontfamily{pcr}\selectfont .doc} files, and some {\fontfamily{pcr}\selectfont .pdf} files were misidentified as {\fontfamily{pcr}\selectfont .jpg} files.  The standalone MLP network was nearly as accurate, reaching 96.15\%.

\begin{figure}[]
	\centering  
	\subfloat[\label{fig:SGAN_50}]{\includegraphics[width=0.4\textwidth]{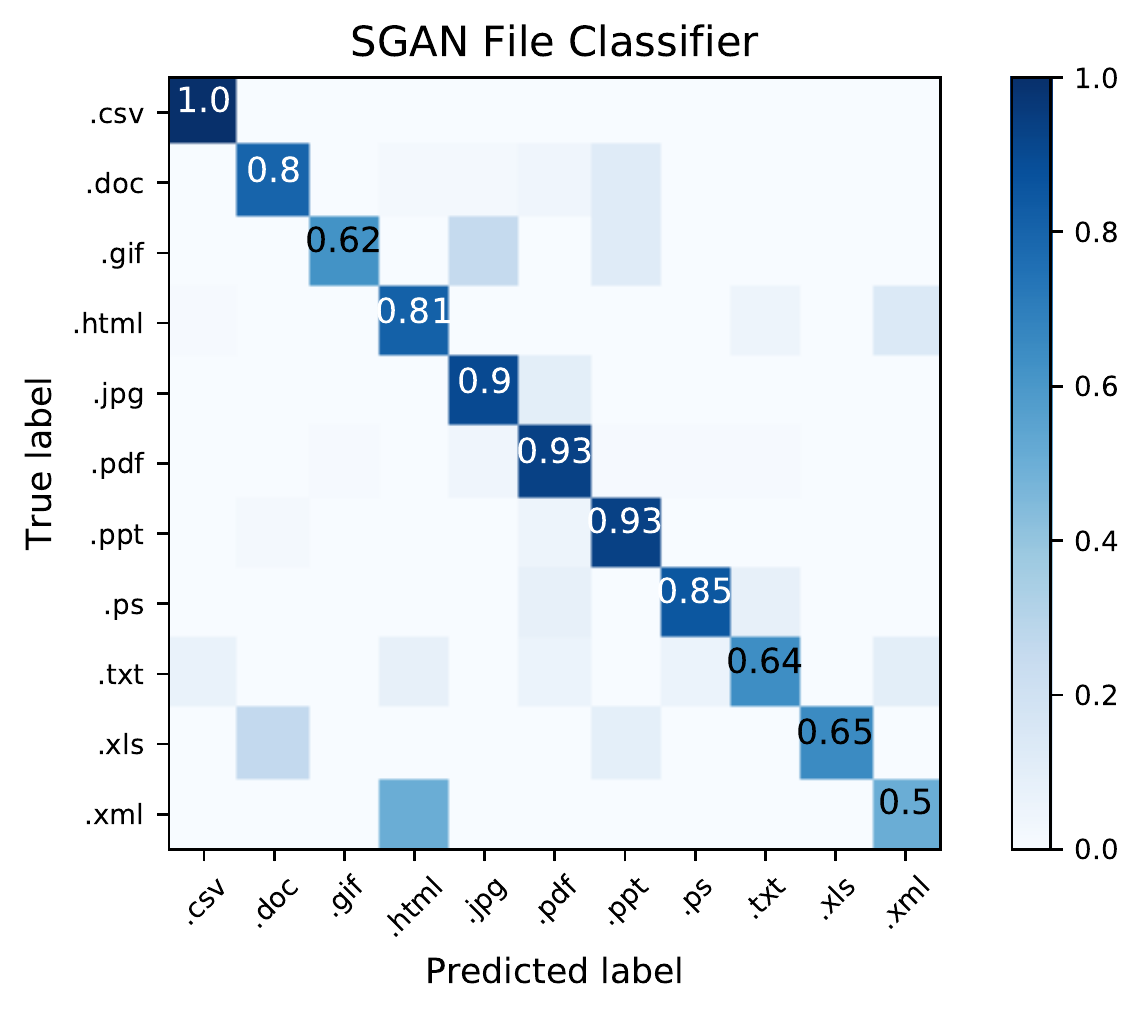}}\par\bigskip
	\subfloat[\label{fig:MLP_50}]{\includegraphics[width=0.4\textwidth]{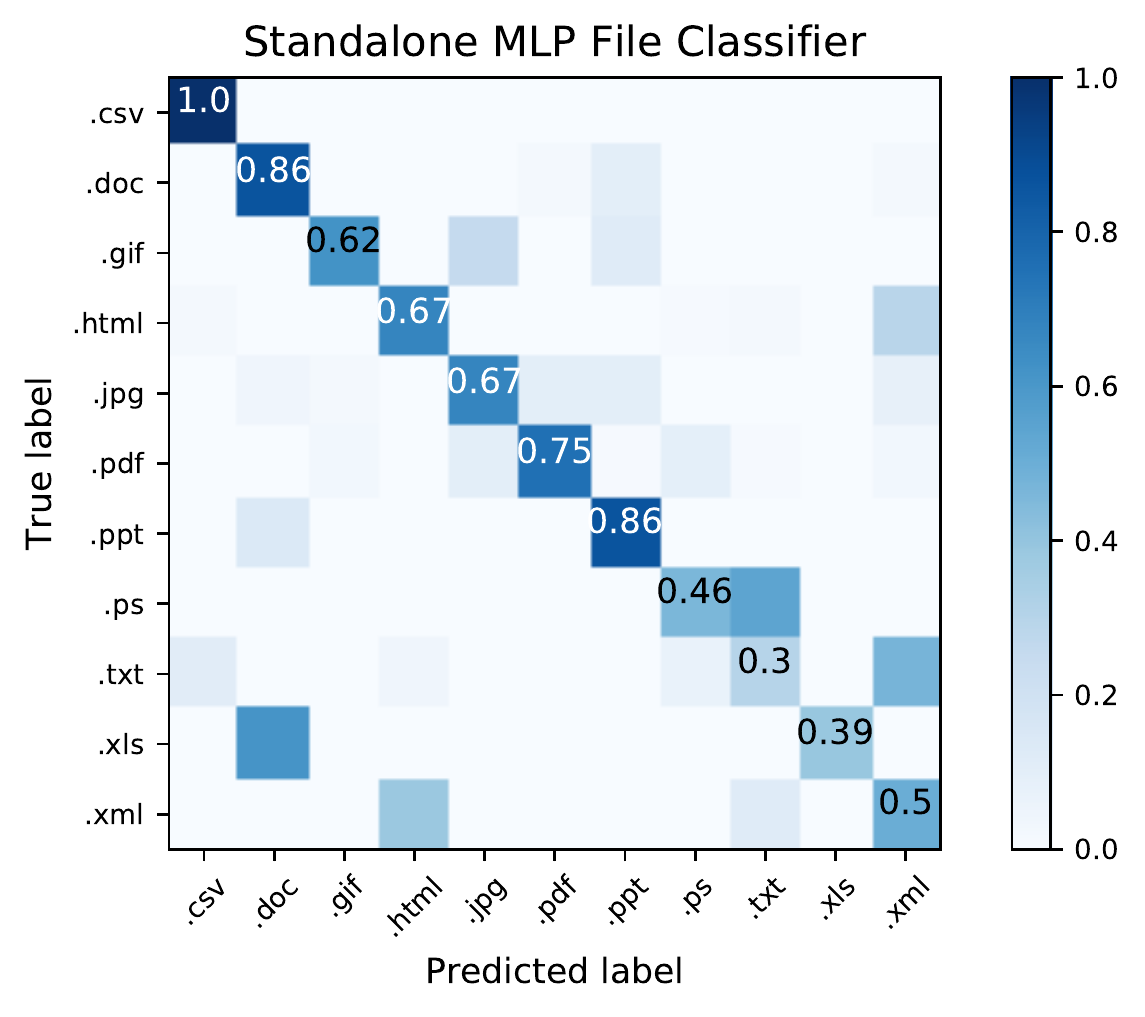}}
	\caption{Confusion matrices for (a)~SGAN and (b)~standalone~MLP trained with 50 supervised samples.}.
	\label{fig:SGAN_MLP_50}
\end{figure}

If we reduce the number of supervised samples provided to our machine learning algorithms, we expect our testing accuracy will be somewhat reduced.  In the course of a forensic investigation, subject matter expertise and a finite amount of time must be prioritized, and since creating a fully-labeled dataset is resource intensive, a worthy goal might be to balance diminishing returns from further training a machine learning algorithm against the time requirements needed for other tasks.  When drastically reducing the training input down to a sample size of 50, only 2.2\% of the training dataset, the SGAN achieved 81.99\% accuracy while the standalone MLP dropped to 62.06\% accuracy.  The confusion matrices for this case are shown in Figure~\ref{fig:SGAN_MLP_50}.  

Comparing Figure~\ref{fig:SGAN_50} and Figure~\ref{fig:MLP_50}, we see that with fewer samples, both neural networks continued to struggle in categorizing {\fontfamily{pcr}\selectfont .xml} and {\fontfamily{pcr}\selectfont .gif} files.  However, with 50 supervised samples, both the SGAN and standalone MLP had more confusion between {\fontfamily{pcr}\selectfont .xml} files and {\fontfamily{pcr}\selectfont .html} files.  The {\fontfamily{pcr}\selectfont .gif} files were incorrectly predicted to be {\fontfamily{pcr}\selectfont .ppt} files as before, but also as {\fontfamily{pcr}\selectfont .jpg} files.  We also see {\fontfamily{pcr}\selectfont .xls} files were incorrectly categorized as {\fontfamily{pcr}\selectfont .doc} files, which is notable as they are both Microsoft products and share the same header information.

The decision tree, XGBoost, and kNN algorithms performed relatively poorly with respect to classification accuracy compared to the neural networks, especially as the number of supervised samples were reduced.  This is likely due to the number of dimensions under assessment with our training and testing samples.  The ``curse of dimensionality'' \cite{murphy_probabilistic_2022} can sometimes be overcome with enough samples, so reducing a training dataset has a predictably deleterious effect on performance.

The fully supervised SGAN model is implemented on GitHub (\url{https://ksaintg.github.io/SGAN-File-Classier/}). Researchers can make use of this implementation to test how altering file headers, changing byte values, deleting portions of files, etc. will affect classification accuracy.  To determine if the file headers would change the SGAN accuracy, we tested our fully-supervised SGAN with a test dataset with altered file headers.  Except for {\fontfamily{pcr}\selectfont .xml}, {\fontfamily{pcr}\selectfont .html}, and {\fontfamily{pcr}\selectfont .txt} files which do not make use of file headers, we replaced the first six bytes of each test dataset file with the hexadecimal string {\fontfamily{pcr}\selectfont AA BB CC DD EE FF}.  The test accuracy for determining the file type between files with altered and unaltered file headers was nearly identical.  Training for up to 300 epochs on a fully-supervised training dataset with unaltered headers, there was only a disparity in altered vs. unaltered test accuracy at epoch 135, where the difference was 0.14\%.

\section{Conclusion and future work}\label{conclusion}
The adversarial training of a neural network produced encouraging results in terms of classification accuracy.  While the neural networks were more complex to train than the other machine learning algorithms, the accuracy results were far superior.  Though the SGAN was the most complex of all the models, its accuracy was the best at correctly classifying files based on their byte value distribution, especially with few supervised samples.  Once trained, the time difference in classifying the dataset between any of the algorithms was indistinguishable.  This work leads to future research using additional neural network architectures and using our spoofed histograms from the generator network to improve other machine learning algorithms.

\bibliographystyle{IEEEtran}
\bibliography{file_classify}

\end{document}